\documentclass[a4paper, twocolumn]{article}

\usepackage[utf8]{inputenc}
\usepackage[T1]{fontenc}

\usepackage[a4paper,
            top=1in,bottom=1in,
            left=1in,right=1in,
            footskip=30pt]{geometry}

\usepackage{microtype}

\usepackage{graphicx}
\usepackage{amsmath,amssymb}
\usepackage{subcaption}
\usepackage{url}
\usepackage{parskip}

\usepackage{fancyhdr}
\fancypagestyle{plain}{
  \fancyhf{}%
  \fancyfoot[C]{Page \thepage}%
}
\title{\bfseries MoE-Beyond: Learning-Based Expert Activation Prediction on Edge Devices}

\author{%
  \begin{tabular}{@{}cc@{\hskip 2em}c@{}}
    \shortstack{%
      \bfseries Nishant Gavhane\\
      Univ.\ of Pennsylvania\\
      \texttt{ngavhane@seas.upenn.edu}
    }
    &
    \shortstack{%
      \bfseries Arush Mehrotra\\
      Univ.\ of Pennsylvania\\
      \texttt{arushm@seas.upenn.edu}
    }
    \\[2ex]
    \shortstack{%
      \bfseries Rohit Chawla\\
      Univ.\ of Pennsylvania\\
      \texttt{rchawla@seas.upenn.edu}
    }
    &
    \shortstack{%
      \bfseries Peter Proença\\
      Univ.\ of Pennsylvania\\
      \texttt{peterpro@seas.upenn.edu}
    }
  \end{tabular}%
}

\date{}

\begin{document}

% two-column-friendly title+abstract:
  \maketitle
    \begin{abstract}
    \textit{The deployment of large-scale Mixture-of-Experts (MoE) models on edge devices presents significant challenges due to memory constraints. While MoE architectures enable efficient utilization of computational resources by activating only a subset of experts per inference, they require careful memory management to operate efficiently in resource-constrained environments. Traditional heuristic-based expert caching strategies such as MoE-Infinity struggle to maintain high cache hit rates as models parameters scale. In this work, we introduce MoE-Beyond, a learning-based expert activation predictor trained to predict expert activations during autoregressive decoding. By framing the task as a multi-label sequence prediction problem, we train a lightweight transformer model on 66 million expert activation traces extracted from LDJnr-Puffin dataset~\cite{LDJnrPuffin} using DeepSeek-V2-Chat-Lite MoE. Our predictor generalizes effectively across unseen prompts from WebGLM-QA dataset \cite{THUDMWebGLMQA}, achieving 97.5\% accuracy and an 86.6\% F1-score. Simulation results show that MoE-Beyond improves GPU cache hit rate from 17\% to 72\% when only 10\% of experts fit in GPU cache, outperforming heuristic baselines.} \url{https://github.com/ngavhane/moe-beyond}
\end{abstract}

\section{Introduction}

Mixture-of-Experts (MoE) models have emerged as a compelling example for scaling deep learning architectures by partitioning large neural networks into multiple specialized “experts” and activating only a small subset at inference time. Early MoE designs demonstrated the feasibility of this approach in machine translation and language modeling, trading off dense parameter counts for conditional computation to achieve higher model capacity without proportional increases in runtime cost. Over the past several years, MoE architectures have evolved from modest-scale research prototypes into the backbone of state-of-the-art large language models, such as Switch-Transformers and DeepSeek, enabling models with hundreds of billions of parameters while keeping per-token compute manageable.

\noindent Despite their theoretical efficiency, MoE models introduce new system challenges related to expert management and memory. In traditional server deployments—where high-end multi‑GPU clusters are available—MoE serving frameworks typically rely on reactive, cache‑based offloading: experts are fetched on demand into GPU memory using least‑recently‑used or static caching policies, and evicted when capacity is exceeded. While effective when model widths and traffic predictability are moderate, these heuristics struggle as expert counts grow and access patterns become less skewed, yielding poor cache hit rates and high PCIe transfer overheads.

In contrast, edge deployments present an entirely different set of constraints. Consumer‑grade GPUs and integrated accelerators often have only tens of gigabytes of memory, far below the hundreds required by modern MoE models. Moreover, edge systems generally operate on batch size one, processing single prompts sequentially rather than serving large batched workloads. This shift amplifies locality in expert activation patterns within a single request but eliminates cross‑request reuse, rendering server‑style caching policies suboptimal. Edge environments therefore demand MoE inference strategies that: (i) minimize memory footprint by selectively offloading experts to host memory, (ii) predict future expert usage to prefetch critical parameters before they stall computation, and (iii) exploit the skew in within‐request expert activations while avoiding the overhead of large‑scale cache management.

Based on these requirements, we propose MoE-Beyond, a learning-based expert activation predictor to frame expert activation prediction during MoE model inference as a multi-label sequence prediction problem. Our work extends the core innovation of MoE-Infinity ~\cite{moe_infinity} by leveraging its Sparsity-Aware Expert Cache, which is designed based on trace analysis insights to optimize Mixture-of-Experts (MoE) inference performance for a batch size of 1. Our contributions are as follows:

\textbf{Contribution 1:} We independently reproduced MoE-Infinity's core insight regarding sparsity in expert activation patterns for MoE inference with batch size of 1. To validate this, we analyzed 122 prompts from the LDJnr-Puffin dataset~\cite{LDJnrPuffin} using the DeepSeek-V2-Lite MoE model with NVIDIA A100.

\textbf{Contribution 2:} We created a dataset of approximately ~100 million expert activation trace points by inferencing 6994 prompts from the LDJnr-Puffin dataset using the DeepSeek-V2-Lite MoE model and recording detailed information for each generated token, such as layer ID, batch number, token value, activated expert IDs, and token embedding vectors.

\textbf{Contribution 3:} We developed a light-weight transformer-based expert activation predictor to outperform heuristic methods like MoE-Infinity’s cosine similarity approach.

MoE-Beyond has been open-sourced on GitHub, including the complete codebase, datasets, and pretrained model weights.

\section{Background}

\subsection{MoE}

Large Language Models (LLMs) have changed natural language processing by using deep transformer architectures that leverage self-attention to understand long-range dependencies. Dense LLMs, such as GPT-3 and BERT, use all model parameters for each token during inference, which leads to a direct and linear relationship between model size and computational cost. While this dense activation simplifies the implementation, it has probems when scaling. It results in billions of parameters, and as a result, inference latency and memory requirements escalate proportionally, which limits deployments to powerful servers.

Mixture-of-Experts (MoE) introduces sparsity into this paradigm by interleaving expert layers among standard transformer blocks. The key idea behind a "mixture" is that each token’s representation is formed by combining the outputs of multiple expert subnetworks, weighted by scores generated from a gating mechanism. In a dense MoE configuration, the gating network produces a full probability distribution over the entire expert pool, and every expert contributes to the final representation through a weighted sum. This soft gating implements an ensemble of experts—each expert can specialize on different patterns or subdomains—while allowing the model to learn how much each expert should influence the output. Although this mixture enriches expressivity, it still incurs computation and memory costs proportional to the number of experts, akin to a wider dense layer.

In contrast, sparse MoE models employ hard gating, where a lightweight router network evaluates each expert for a given token and activates only the top-k experts. Models such as DeepSeek-V2-Lite and Switch Transformers are examples of sparse MoEs. This gating mechanism transforms the mixture into a conditional selection process. It enables selective computation, where only the chosen k experts perform inference, reducing the need for every expert to be active and thus significantly cutting down computational work. Additionally, it ensures efficient memory use, as activating a subset of experts lowers peak memory footprint and bandwidth demands, with unused experts remaining inactive. The mechanism also supports adaptivity, allowing the model to dynamically adjust computational effort by varying the value of k or the gating threshold, tailoring resource usage to the complexity of the task at hand.

\subsection{Expert Activation Sparsity}

Our research validates the key insights from the MoE-Infinity framework~\cite{moe_infinity} regarding expert activation patterns in Mixture-of-Experts (MoE) models operating in single-user environments. Xue et al.\ demonstrated that with a batch size of one, typical in edge inference scenarios, MoE models exhibit significant activation sparsity. Their analysis showed that for models with approximately 100 experts, only 5\% of experts are activated per token, while smaller models like Mixtral-8x7B (8 experts/layer) activate around 25\% of experts per token. 

\begin{figure*}[t]
    \centering
    \includegraphics[width=\textwidth]{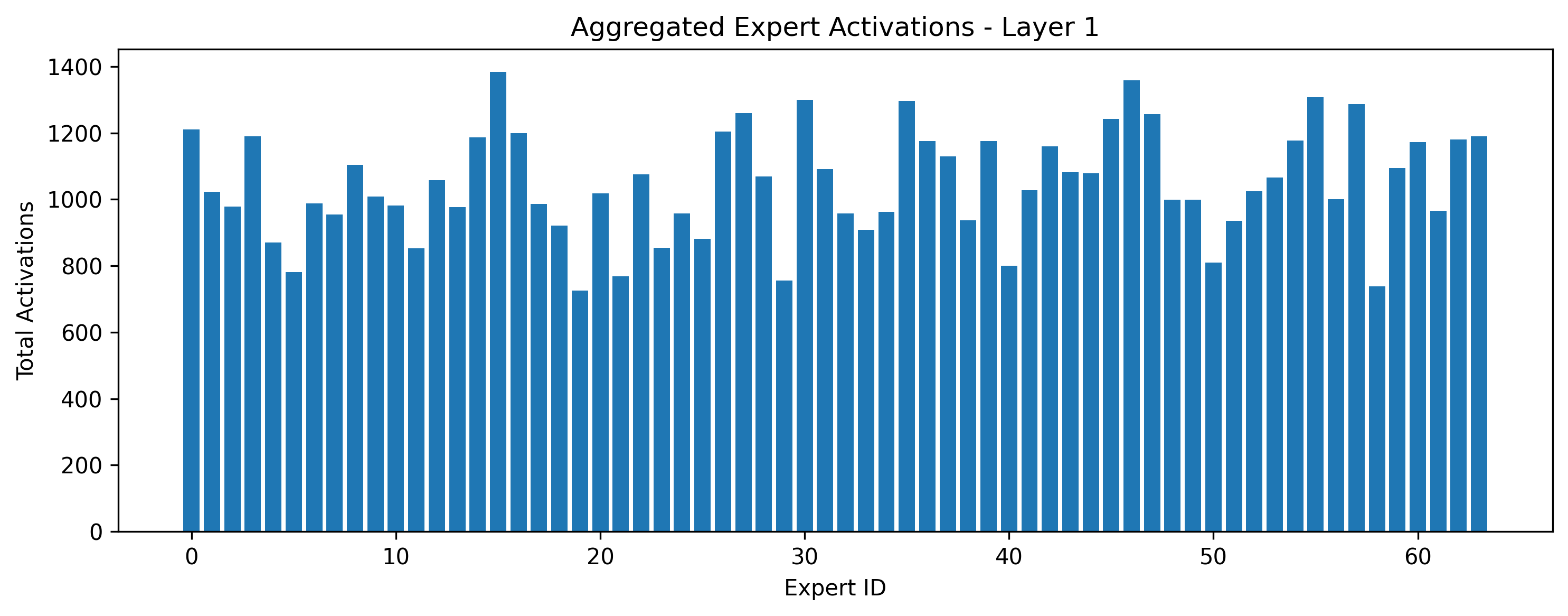}
    \caption{Aggregated expert activations for Layer 1 across 122 prompts.}
    \label{fig:layer1_activations}
\end{figure*}

We reproduced these findings using 122 prompts from the Puffin dataset~\cite{LDJnrPuffin}, a collection of multi-turn conversations generated with GPT-4. Our trace analysis with DeepSeek-V2-Lite (27 layers, 64 experts per layer) produced three key visualizations that substantiate the original paper's claims. Figure~\ref{fig:layer1_activations} presents the aggregated expert activations for Layer 1 across 122 prompts, revealing a uniform distribution, each expert receiving between 800--1400 activations, confirming that expert popularity becomes evenly distributed when observed across multiple requests. 

\begin{figure*}[t]
    \centering
    \includegraphics[width=0.8\textwidth]{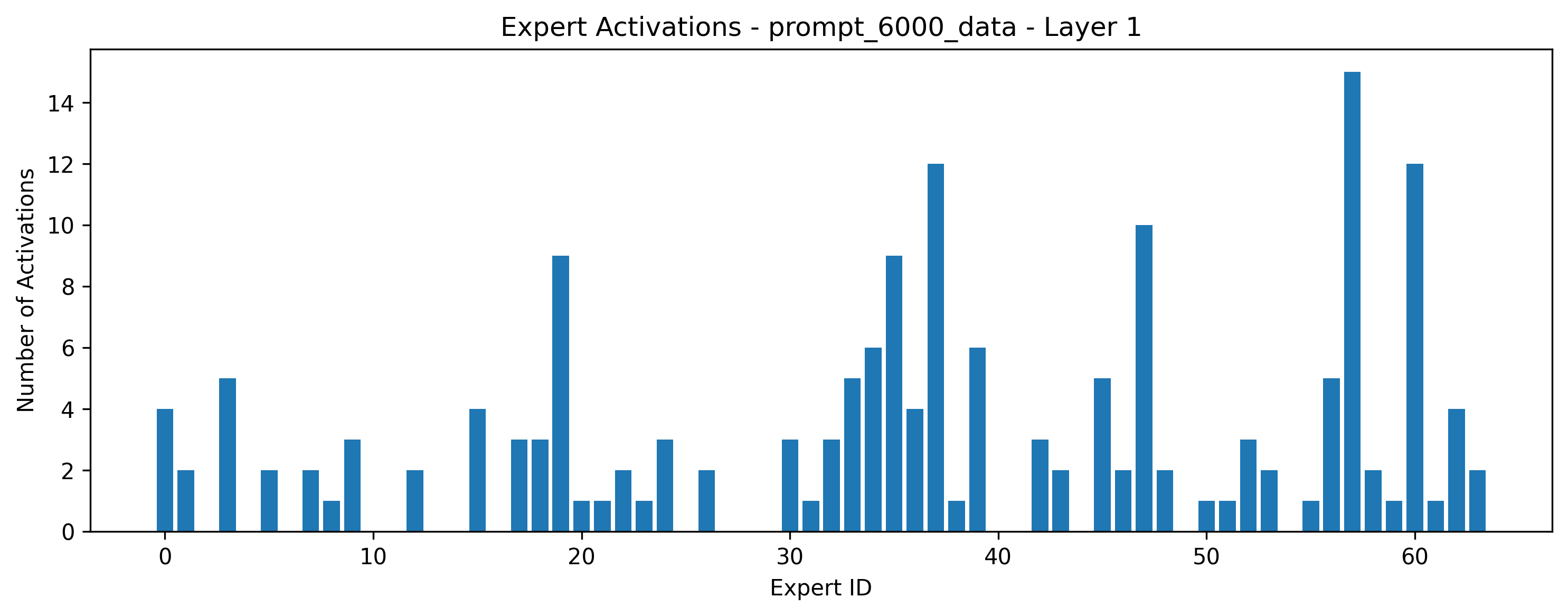}
    \caption{Expert activations for a single prompt (\#6000), showing sparse usage.}
    \label{fig:single_prompt_activations}
\end{figure*}

\begin{figure*}[t]
    \centering
    \includegraphics[width=\textwidth]{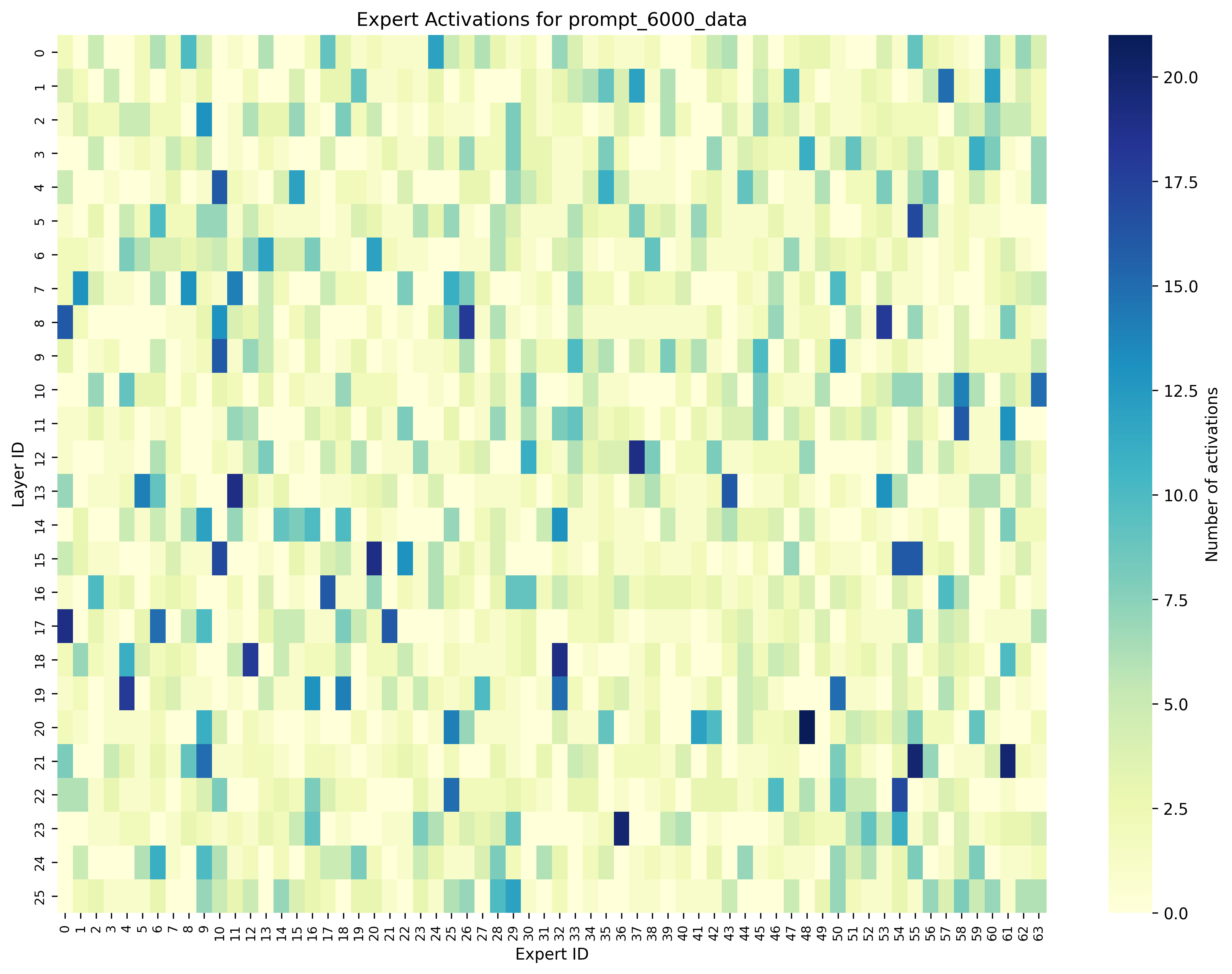}
    \caption{Layer-wise expert activation heatmap across 27 layers of DeepSeek-V2-Lite MoE. Highlighted regions indicate consistent expert reuse.}
    \label{fig:layerwise_heatmap}
\end{figure*}

In contrast, Figure~\ref{fig:single_prompt_activations} displays activations for a single prompt (prompt \#6000), showing dramatic sparsity with only a small subset of experts (those with peaks at expert IDs $\sim$19, 35, 37, 47, 57, and 60) receiving significant activations. The comprehensive heatmap in Figure~\ref{fig:layerwise_heatmap} extends this analysis across all 27 layers, demonstrating consistent expert reuse patterns within a single prompt context. The stark contrast between the uniform multi-prompt distribution and the highly skewed single-prompt pattern provides empirical evidence supporting MoE-Infinity's approach of request-aware expert caching over traditional frequency-based methods. These findings highlight the importance of tracing activation patterns at the request level for efficient MoE model serving.


\subsection{GPU VRAM as an Expert Cache}

For MoE models deployed on edge devices, efficient memory management is a core concern due to limited hardware resources. MoE architectures are inherently sparse and only a subset of experts is activated per inference. This makes them amenable to dynamic memory strategies. In this context, GPU VRAM acts as a high-speed cache, storing active or frequently used experts to reduce latency and memory overhead.

Storing all experts in GPU VRAM is typically infeasible on edge devices, where VRAM is limited and expensive in terms of energy and footprint. Instead, caching strategies such as Least Recently Used (LRU) or Least Frequently Used (LFU) are employed to retain only the most relevant experts based on recent activation patterns. This allows the system to prioritize memory resources for computation-critical data and minimize expert load times from slower storage hierarchies \cite{moe_infinity}. 

The primary benefit of this approach is reduced inference latency. By keeping hot experts resident in VRAM, the system avoids costly memory transfers from host RAM or flash storage. Furthermore, it enhances throughput by minimizing data movement, which is especially valuable in real-time applications on the edge. 

However, there are notable challenges. Cache management introduces overhead in tracking expert usage and orchestrating eviction policies. Moreover, if expert access patterns are highly dynamic or unpredictable, cache miss rates may increase, negating performance gains. Prefetching techniques and predictive caching, such as those introduced in ProMoE, aim to alleviate this by forecasting future expert activations based on intermediate computation results \cite{promoe}. 

Overall, treating GPU VRAM as an expert cache is a pragmatic solution to the memory limitations of edge inference, balancing resource constraints with the computational advantages of MoE models.

\section{Methodology}

\subsection{Heuristics Based Approach}
\label{sec:heuristics}

Early MoE serving systems leaned on simple, rule‑based predictors. DeepSpeed‑MoE \cite{deepseed_moe}, for example, eagerly loads every expert in the next layer, assuming dense‑model locality; this works for transformers but over‑fetches badly once a sparse router is introduced. A more global variant, BrainStorm, counts how often each expert fires across an entire workload and retains the “popular” ones, but once many prompts are merged these counts flatten out and the hit‑rate collapses \cite{brainstorm}. 

MoE-Infinity exploits the observation that, on single-user machines, an MoE LLM typically activates only a handful of experts during the autoregressive decode phase. It constructs a request-level Expert Activation Matrix (EAM): after each token is routed, the system logs a sparse bit-vector that marks which experts fired in that layer. A small “EAM-collection” keeps the most recent sketches; when the cache must decide what to prefetch, it finds the sketch whose cosine distance to the current one is smallest and brings the entire matched expert group into GPU memory. Because prompts tend to revisit the same semantic sub-skills, this strategy achieves 2.7 - 13.7x performance improvements over strong baselines \cite{moe_infinity}.

A key limitation is that EAM matching relies on similarity to recent activation sketches. The dense bit-vector grows linearly with the number of experts, so both memory footprint and cosine-search cost rise for very wide MoEs, and the algorithm assumes a single-request decode loop—interleaved streams intermingle sketches and dilute the signal unless the history window is greatly enlarged. Moreover, because eviction is based only on past reuse scores, it ignores cross-layer gate correlations discovered in later work, occasionally evicting an expert just before a downstream layer needs it.

\subsection{Learning Based Approach}

\subsubsection{Problem Formulation and Input Representation}

We formulate expert activation prediction as a multi-label sequence modeling task where the model receives token embeddings $X \in \mathbb{R}^{T \times 2048}$ and layer position identifiers $L \in \mathbb{Z}^T$ as input. The output is a binary vector $y_t \in \{0, 1\}^{64}$ at each sequence position $t$, indicating which of the 64 experts should activate. To capture layer-specific dynamics, we embed layer IDs using a learned embedding layer $E_{\text{layer}} \in \mathbb{R}^{27 \times 512}$, concatenating these with token embeddings to form composite features $F \in \mathbb{R}^{T \times 2560}$. The maximum sequence length is constrained to 512 tokens through truncation and smart padding, balancing computational efficiency with context preservation.

\subsubsection{Transformer-Based Prediction Architecture}

The core architecture employs a 4-layer transformer encoder with 8 attention heads and 2048-dimensional feedforward networks. Input features are first projected to a 512-dimensional space through a linear layer before entering the transformer stack. The encoder utilizes masked self-attention to prevent information leakage between padded and actual tokens, with dropout ($p = 0.1$) applied to attention weights and feedforward outputs for regularization. Final expert logits are generated through a 2-layer MLP head with GELU activation and dimension reduction (512 $\rightarrow$ 64), enabling simultaneous prediction for all 64 experts in the layer.

\subsubsection{Training Protocol and Optimizations}

We implement dynamic mixed-precision training using PyTorch's Automatic Mixed Precision (AMP), scaling gradients with a GradScaler to prevent underflow. The AdamW optimizer ($\beta_1 = 0.9$, $\beta_2 = 0.98$) applies layer-wise learning rate decays: $LR_{\text{input\_proj}} = 10^{-4}$, $LR_{\text{encoder}} = 0.9 \times 10^{-4}$, $LR_{\text{head}} = 0.8 \times 10^{-4}$, with L2 weight decay of 0.01. Gradient norms are clipped at 1.0 to prevent explosion. Training employs a batch size of 4, optimizing performance on datasets with 6000 files containing variable-length sequences. We trained the model for 10 epochs. Early stopping is applied if validation loss does not improve for 3 consecutive epochs.

\subsubsection{Training Validation and Analysis Framework}

We evaluate predictions using two complementary metrics: position-wise accuracy and macro F1 score across all experts.

\begin{itemize}
    \item Position-wise accuracy measures the fraction of tokens for which the predicted set of active experts exactly matches the ground truth. Specifically, at each token position $t$, the model outputs a set of top-6 experts based on the sigmoid-activated logits, where experts are selected if their corresponding activation probability exceeds 0.5. The position-wise accuracy is computed by comparing the predicted set of top-6 experts for each token with the ground truth, which is represented as a multi-hot encoded vector of activated experts. This metric helps assess how accurately the model is selecting the right set of experts for each token in the sequence.
    \item Macro F1 score is computed across all 64 experts, treating each expert's prediction as a binary classification problem (expert activated or not). For each expert, we compute precision (the proportion of predicted activations that are correct) and recall (the proportion of actual activations that are correctly predicted), and then calculate the F1 score as the harmonic mean of precision and recall. The macro F1 score is the average of the F1 scores across all experts, which gives a balanced view of the model's performance across all experts, taking into account both precision and recall.
\end{itemize}

During validation, the model selects the top-6 experts for each token by applying sigmoid activation to the logits. These predictions are compared against the ground truth activations. A custom analysis loop also extracts layer-specific prediction patterns, logging the per-layer expert agreement rates to TensorBoard for visual inspection. This allows for a deeper understanding of how well the model generalizes across different layers and experts. The implementation further optimizes memory usage by caching processed sequences using an LRU strategy (capacity=1000), which accelerates epoch iterations without sacrificing memory efficiency. The best model is selected based on validation loss, ensuring that the model with the lowest loss on the validation set is retained.

\begin{figure*}[h!]
    \centering
    \includegraphics[width=1.0\textwidth]{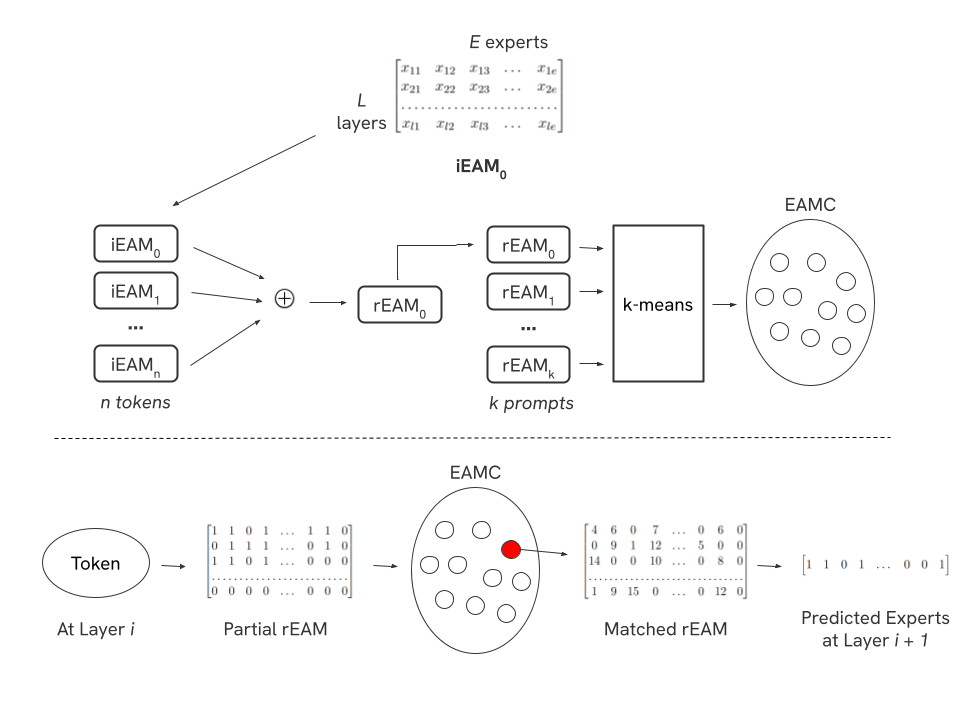}
    \caption{Simulator diagram for EAMC creation (top) and expert prediction once EAMC has been created (bottom). iEAMs are created for each token and then accumulated across an entire prompt to create a rEAM. rEAMs are passed through k-means to create the EAMC. A partial rEAM is created for the first \textit{n} tokens in a test prompt and then passed through the EAMC to find the nearest neighbor. The matched rEAM informs which experts to prefetch for the next layer of the current token.}
    \label{fig:cache_hit_rate}
\end{figure*}

\subsubsection{Training Dynamics and Convergence}

\begin{figure*}[h!]
    \centering
    \includegraphics[width=0.9\textwidth]{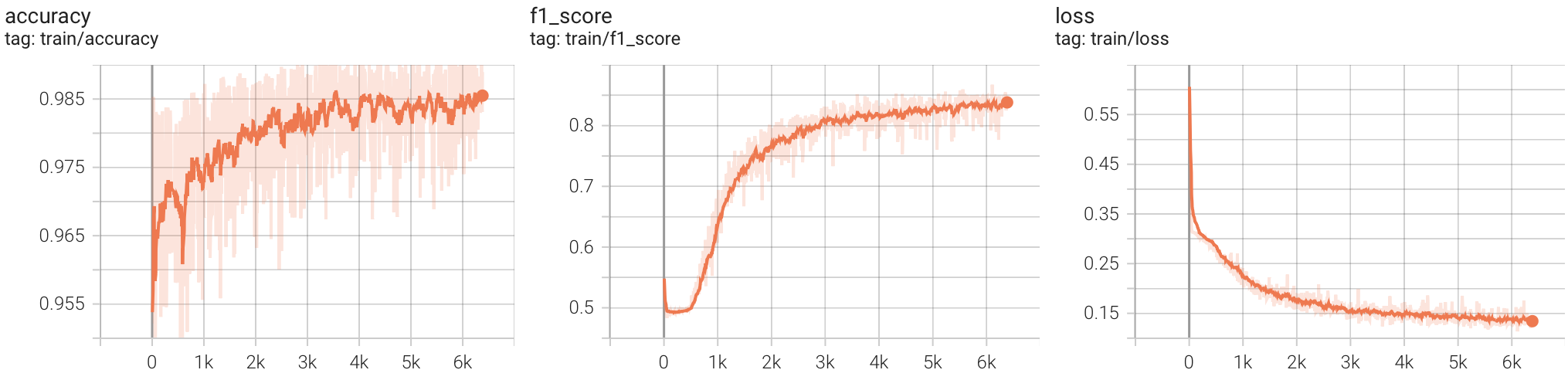}
    \caption{Training metrics across training steps. (a) Accuracy, (b) F1-score (c) Loss.}
    \label{fig:training_metrics}
\end{figure*}

Our transformer-based expert predictor exhibited strong learning dynamics across the training phase, as illustrated in Figure~\ref{fig:training_metrics}. The training accuracy (Figure~\ref{fig:training_metrics}a) demonstrated consistent improvement from an initial 96\% to 98.9\% (98.4\% smoothed) over 6382 training steps, with the high starting value reflecting the inherent 6:58 class imbalance in expert activation patterns. This class imbalance highlights why accuracy alone is insufficient as an evaluation metric. The training loss curve (Figure~\ref{fig:training_metrics}c) shows steady convergence from 0.35 to 0.131, with the steepest decline occurring during the first 2000 steps. Most notably, the training F1-score (Figure~\ref{fig:training_metrics}b) exhibits significant improvement from approximately 0.5 to 0.86, indicating the model's growing ability to correctly predict both active and inactive experts despite the class imbalance. The training process required 2 days on our hardware configuration with diminishing returns observed after approximately 4000 steps.

\subsubsection{Validation Performance and Generalization}

\begin{figure*}[h!]
    \centering
    \includegraphics[width=0.9\textwidth]{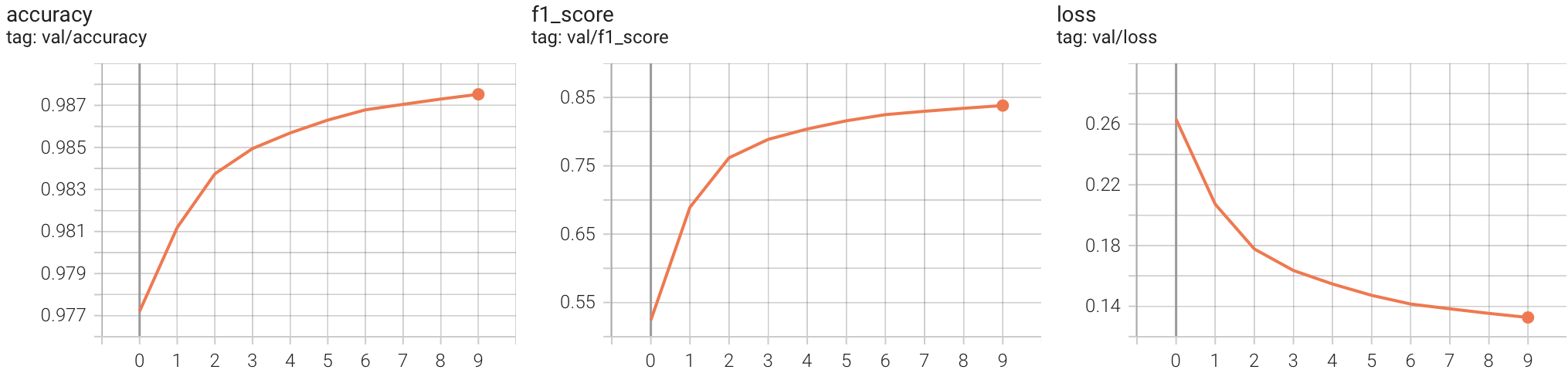}
    \caption{Validation metrics across epochs (a) Accuracy, (b) F1-score (c) Loss.}
    \label{fig:validation_metrics}
\end{figure*}

The validation metrics, shown in Figure~\ref{fig:validation_metrics}, demonstrate the model's strong generalization capabilities to unseen data. The validation accuracy (Figure~\ref{fig:validation_metrics}a) stabilized at 98.7\%, closely tracking the training accuracy trend with minimal overfitting. The validation loss (Figure~\ref{fig:validation_metrics}c) similarly decreased from approximately 0.25 to 0.133, maintaining a small gap with training loss throughout the process. Most significantly, the validation F1-score (Figure~\ref{fig:validation_metrics}b) reached 0.85 after starting from approximately 0.55, nearly matching the training F1-score of 0.86. This minimal difference between training and validation F1-scores (0.86 vs. 0.85) confirms the model's robust generalization capability while predicting expert activation patterns across diverse contextual inputs. The consistent performance across both training and validation metrics suggests the model architecture strikes an appropriate balance between capacity and generalizability for the expert activation prediction task.

\section{Evaluation}

\subsection{Experimental Methodology}

\subsubsection{Models}

To evaluate MoE-Beyond, we consider the DeepSeek-V2-Lite model from Huggingface Hub. DeepSeek-V2-Lite is an open-source, MoE language model in which each MoE layer comprises of 2 shared experts and 64 routed experts, with 6 experts activated per token. This model consists of 15.7 billion total parameters and activates 2.4 billion parameters with each forward pass. 

\subsubsection{Datasets}

To train and evaluate MoE-Beyond, we curated two datasets by extracting expert activation traces from the DeepSeek-V2-Lite model.

\paragraph{\textbf{Training Dataset}}
We utilized the Puffin dataset \cite{LDJnrPuffin}, which comprises 3,000 high-quality, multi-turn conversations generated using GPT-4. Each conversation averages over 1,000 tokens and includes diverse topics such as physics, biology, math, and chemistry. We processed 6,994 prompts from this dataset by running them through the DeepSeek-V2-Lite model. For each token in the generated sequences, we recorded the following information:

\begin{itemize}
    \item \textbf{Layer ID}: The specific layer within the model.
    \item \textbf{Batch Number}: Identifier for the batch during processing.
    \item \textbf{Token}: The token being processed.
    \item \textbf{Activated Expert IDs}: The IDs of the experts activated for the token.
    \item \textbf{Token Embedding Vector}: The embedding representation of the token.
\end{itemize}

This process resulted in approximately 66 million training samples, which were used to train our transformer-based expert activation predictor.

\paragraph{\textbf{Test Dataset}}

For evaluation, we selected 100 prompts from the WebGLM-QA dataset \cite{THUDMWebGLMQA}, a collection designed for web-enhanced question-answering tasks. Similar to the training data, we processed these prompts through the DeepSeek-V2-Lite model and extracted the same set of features per token. The predicted expert activations from MoE-Beyond were then compared against the actual activations to assess performance.

\subsubsection{Baseline}

We evaluate MoE-Beyond against MoE-Infinity \cite{moe_infinity}, a state-of-the-art offloading-efficient serving system for sparse MoE models. As described in Section~\ref{sec:heuristics}, MoE-Infinity constructs request-level Expert Activation Matrices (EAMs) and uses heuristic cosine similarity to predict and prefetch required experts into GPU memory. While MoE-Infinity achieves strong cache hit rates and latency reduction, its performance degrades under domain shifts or interleaved inputs due to reliance on recent activation sketches. 

MoE-Beyond builds on this foundation by introducing a transformer-based expert activation predictor trained on large-scale trace data. This learning-based method improves generalization and anticipates expert usage more accurately.

\subsubsection{Simulator}

The evaluation for MoE-Infinity is performed with a trace‑based simulator. For every prompt we run DeepSeek‑V2-Lite once and log, to a CSV file, each Layer ID together with the list of Activated Expert IDs.  
A single prompt trace is then folded into a request‑level Expert Activation Matrix (REAM): an $L{\times}E$ histogram over the model’s 27 layers and 64 experts. The first $N$ traces are stored in an EAM‑Collection (EAMC).

During simulation each test prompt is replayed token by token.  
The first \(n\) tokens simply warm an LRU Expert Cache so that both cache and EAMC start in realistic states. From token \(n{+}1\) onward we (i)~flatten the partial REAM accumulated so far, (ii)~compute its cosine similarity to every sketch in the EAMC, and (iii)~select the most similar sketch to predict which experts will fire in the next layer. These predicted experts are prefetched into Expert Cache; the simulator then reveals the ground‑truth expert IDs from the trace. A prediction hit is recorded if the ground‑truth expert appears in the predicted set, and a cache hit if it is already resident. Sweeping the cache size and aggregating hits across hundreds of prompts yields prediction accuracy and cache‑hit rate, allowing us to quantify how effectively the MoE‑Infinity heuristic turns trace information into real‑time memory savings.

\subsection{Results}

In this section, we evaluate MoE-Beyond against MoE-Infinity across two core metrics: GPU Cache Hit Rate and Transformer Evaluation Accuracy. These metrics reflect both the system-level and model-level performance gains that result from more accurate expert activation forecasting.

\subsubsection{GPU Cache Hit Rate}

Cache hit rate serves as our primary system-level evaluation metric, capturing the proportion of expert activations successfully served from the GPU cache without offloading. This is particularly critical on edge devices and memory-constrained GPUs, where minimizing PCIe transfer latency and maximizing in-memory utilization is key. 

Figure~\ref{fig:cache_hit_rate} shows how MoE-Beyond consistently outperforms MoE-Infinity across varying GPU expert capacities. Even when only 10\% of experts fit in memory, MoE-Beyond achieves a hit rate of over 70\%, compared to just 17\% for MoE-Infinity. As the memory budget increases, MoE-Beyond maintains a lead of 10–25 percentage points, converging to 100\% hit rate faster. 

This improved performance highlights the benefit of using a learned model to anticipate expert reuse patterns, rather than relying on heuristic similarity to recent activation traces. These gains translate to lower latency and reduced bandwidth overhead in real-world deployment scenarios.

\begin{figure*}[h!]
    \centering
    \includegraphics[width=0.8\textwidth]{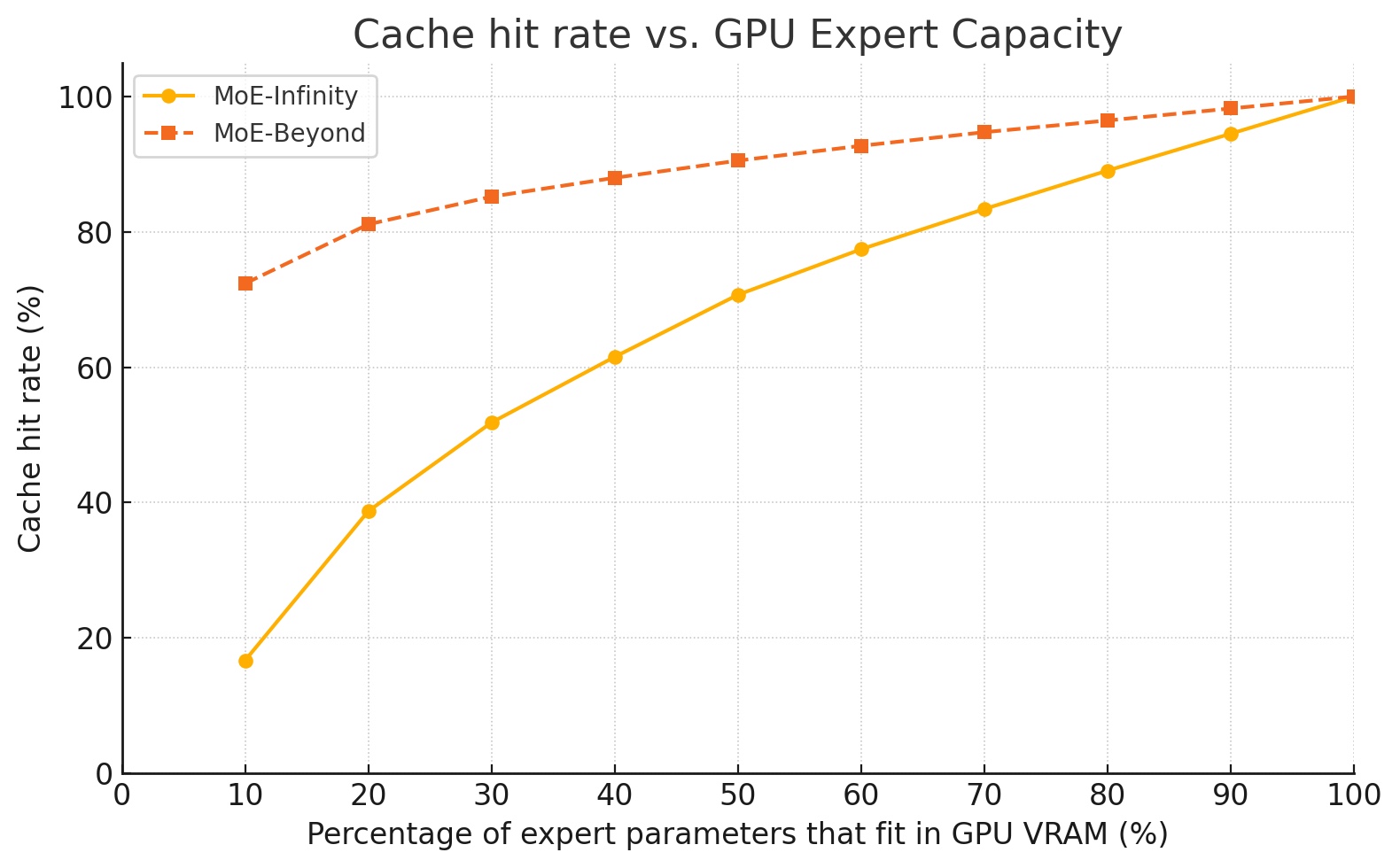}
    \caption{Cache hit rate (\%) vs. GPU expert capacity (\%) for MoE-Infinity and MoE-Beyond.}
    \label{fig:cache_hit_rate}
\end{figure*}

\subsubsection{Transformer Evaluation Metrics}

To assess the standalone performance of our expert activation predictor, we evaluated the trained transformer model on a held-out test set of 100 prompts from the WebGLM-QA dataset. The model achieved the following results:

\begin{table}[h!]
\centering
\begin{tabular}{|c|c|}
\hline
\textbf{Metric} & \textbf{Value} \\
\hline
Accuracy & 97.55\% \\
F1-Score & 86.18\% \\
\hline
\end{tabular}
\caption{Validation accuracy and macro F1-score on the 100-prompt test set.}
\label{tab:transformer_eval}
\end{table}

These results demonstrate that our transformer model generalizes well to unseen inputs, effectively capturing expert activation dynamics necessary for high prediction quality and cache efficiency.

\section{Limitations}

While our learned predictor markedly improves cache hit rate and prediction hit rate over fixed heuristics, it inherits several practical constraints that keep it from being a drop-in replacement in large-scale serving stacks. 

First, the method assumes a batch size of one. Sparse expert activation, the very signal the model exploits, collapses when prompts are merged, because gating decisions from multiple users superpose and drive most experts above threshold. In high-throughput inference, where requests are routinely micro-batched to amortize compute, the learned model loses discriminative power and degenerates toward the ``prefetch-everything" regime it was designed to avoid.

Second, the predictor is tightly coupled to a specific backbone. The token embeddings and layer-ID encoding reflect architectural choices; a new MoE configuration therefore requires full retraining—or at minimum expensive fine-tuning. This contrasts with rule-based schemes, such as LRU caching or EAM matching, which transfer across models with little or no modification. Retraining is non-trivial: even with mixed precision and sequence truncation, our four-layer transformer needs $\sim$48 A100 GPU-hours to converge on DeepSeek-V2-Lite, creating a maintenance burden when model revisions are frequent.

Finally, inference latency remains on the critical path. The current implementation predicts only one layer ahead; DMA transfers overlap with the \emph{preceding} layer's compute but not with deeper layers. Extending the horizon would require either a larger transformer (raising latency and memory) or a hierarchical scheduling policy, both of which we leave for future work.

\section{Conclusions}

We introduced MoE-Beyond, one of the first learning-based expert-activation predictors tailored to memory-constrained, single-user inference on edge GPUs.  By framing expert selection as a multi-label sequence task and training a lightweight four-layer transformer on 66 million expert activation traces, our system lifts prediction hit rate from 17\% to 72\% and boosts GPU cache hit rate by up to 55\% compared with the heuristic state-of-the-art (MoE-Infinity).

While promising, the approach is currently specialized to single batch inference and must be retrained for each MoE backbone, and its one-layer look-ahead leaves residual stalls for long contexts.  Future work will explore hierarchical or cross-layer predictors that generalize across model families, extend the prediction horizon, and operate under micro-batched or multi-tenant workloads.  We believe unifying trace-driven learning with hardware-aware scheduling can unlock even larger mixtures on truly edge-scale devices, bringing the accuracy of billion-parameter MoEs to latency-sensitive applications.


\begin{thebibliography}{1}

\bibitem{moe_infinity}
Xue et al., \textit{MoE-Infinity: Activation-Aware Expert Offloading for Sparse Mixture-of-Experts}, 2024. \url{https://arxiv.org/abs/2401.14361}

\bibitem{deepseed_moe}
Rajbhandari et al., \textit{DeepSpeed-MoE: Advancing Mixture-of-Experts Inference and Training to Power Next-Generation AI Scale}, 2022. \url{https://arxiv.org/abs/2201.05596}

\bibitem{promoe}
Ziqi Zhang et al., \textit{ProMoE: Proactive Mixture-of-Experts with Expert Activation Prediction}, 2023. \url{https://arxiv.org/abs/2305.19414}

\bibitem{brainstorm}
Cui et al., \textit{Optimizing Dynamic Neural Networks with Brainstorm}, 2023. \url{https://www.usenix.org/system/files/osdi23-cui.pdf}

\bibitem{LDJnrPuffin}
LDJnr. \textit{Puffin Dataset}. Hugging Face. Available at: \url{https://huggingface.co/datasets/LDJnr/Puffin}

\bibitem{THUDMWebGLMQA}
THUDM. \textit{WebGLM-QA Dataset}. Hugging Face. Available at: \url{https://huggingface.co/datasets/THUDM/webglm-qa}

\end{thebibliography}
\end{document}